\documentclass{article}

\usepackage[final]{nips_2016}

\usepackage[utf8]{inputenc}
\usepackage{hyperref}
\usepackage{url}
\usepackage[dvipsnames]{xcolor}
\usepackage{graphicx}
\usepackage{paralist}
\usepackage{amssymb}
\usepackage{amsmath}
\usepackage{wrapfig}
\usepackage{natbib}
\usepackage[percent]{overpic}

\usepackage{caption, subcaption}

\usepackage{algorithm}
\usepackage{algorithmic}

\newcommand{\myvecsym}[1]{\boldsymbol{#1}}

\newcommand{\vtheta}{\myvecsym{\theta}}

\newcommand{\vTheta}{\myvecsym{\Theta}}

\newcommand{\be}{\begin{equation}}
\newcommand{\ee}{\end{equation}}
\newcommand{\bea}{\begin{eqnarray}}
\newcommand{\eea}{\end{eqnarray}}
\newcommand{\beaa}{\begin{eqnarray*}}
\newcommand{\eeaa}{\end{eqnarray*}}

\DeclareMathAlphabet{\mathpzc}{OT1}{pzc}{m}{n}

\title{Bayesian Optimization in AlphaGo}
\author{Yutian Chen, Aja Huang, Ziyu Wang, Ioannis Antonoglou, Julian Schrittwieser, \\ {\bf David Silver \& Nando de Freitas}\\
  \\ DeepMind, London, UK \\
  \texttt{yutianc@google.com} \\
}

\begin{document}

\maketitle

\begin{abstract}
During the development of AlphaGo, its many hyper-parameters were tuned with Bayesian optimization multiple times. This automatic tuning process resulted in substantial improvements in playing strength. For example,
prior to the match with Lee Sedol, we tuned the latest AlphaGo agent and this improved its 
win-rate from 50\% to 66.5\% in self-play games. This tuned version was deployed in the final match. Of course, since we tuned AlphaGo many times during its development cycle, the compounded contribution was even higher than this percentage.
It is our hope that this brief case study will be of interest to Go fans, and also provide Bayesian optimization practitioners with some insights and inspiration.
\end{abstract}

\section{Introduction}

Bayesian optimization was used as a routine service to adjust the hyper-parameters of AlphaGo  \citep{Silver:2016}  during its design and development cycle, resulting in progressively stronger agents. In particular, Bayesian optimization was a significant factor in the strength of AlphaGo in the highly publicized match against Lee Sedol. 

AlphaGo may be described in terms of two stages: Neural network training, and game playing with Monte Carlo tree search (MCTS). Each of these stages has many hyper-parameters. We focused on tuning the hyper-parameters associated with game playing. We did so because we had reasonably robust  strategies for tuning the neural networks, but less human knowledge on how to tune AlphaGo during game playing.

We meta-optimized many components of AlphaGo. Notably, we tuned the MCTS hyper-parameters, including the ones governing the UCT exploration formula, node-expansion thresholds, several hyper-parameters associated with the distributed implementation of MCTS, and the hyper-parameters of the formula for choosing between fast roll-outs and value network evaluation per move.  We also tuned the hyper-parameters associated with the evaluation of the policy and value networks, including the softmax annealing temperatures. Finally, we meta-optimized a formula for deciding the search time per move during games. The number of hyper-parameters to tune varied from 3 to 10 depending on a tuning task. The results section of this brief paper will expand on these tasks.

Bayesian optimization not only reduced the time and effort of manual tuning, but also improved the playing strength of AlphaGo by a significant margin. Moreover, it resulted in useful insights on the individual contribution of the various components of AlphaGo, for example shedding light on the value of fast Monte Carlo roll-outs versus value network board evaluation. 

There is no analytically tractable formula relating AlphaGo's win-rate and the value of its hyper-parameters. However, we can easily estimate it via self-play, that is by playing an AlphaGo version $v$ against a baseline version $v_0$ for $N$ games and, subsequently, computing the average win-rate:
\begin{equation}
\bar{p}_{v, v_0} = N_{\mathrm{win}} / N.
\end{equation}
By playing several games with different versions of AlphaGo, we can also adopt the  
 BayesElo algorithm  \citep{coulom2008whole}  to estimate a scalar value indicating the strength of each AlphaGo agent.

Since each Go game has only two outcomes, win or lose, the average win-rate is the sample average of a Bernoulli random variable with true win-rate $p_{v, v_0}$. We can also compute confidence intervals easily for a sample of size $N$. The win-rate $p_{v, v_0}(\vtheta)$, or simply $p(\vtheta)$, is a function of the hyper-parameters $\vtheta$ of $v$ with $v_0$ fixed, but the analytical form of this function is unknown.  

Before applying Bayesian optimization, we attempted to tune the hyper-parameters of AlphaGo one-at-a-time using grid search. Specifically, for every hyper-parameter, we constructed a grid of valid values and ran self-play games between the current version $v$ and a fixed baseline $v_0$. For every value, we ran $1000$ games. The games were played with a fixed 5-second search time per move. It took approximately 20 minutes to play one game. By parallelizing the games with several workers, using 400 GPUs, it took approximately 6.7 hours to estimate the win-rate $p(\vtheta)$ for a single hyper-parameter value. The optimization of 6 hyper-parameters, each taking 5 possible values, would have required 8.3 days.
This high cost motivated us to adopt Bayesian optimization.

\section{Methods}

\begin{figure}[t]
\centering
\includegraphics[width=\textwidth]{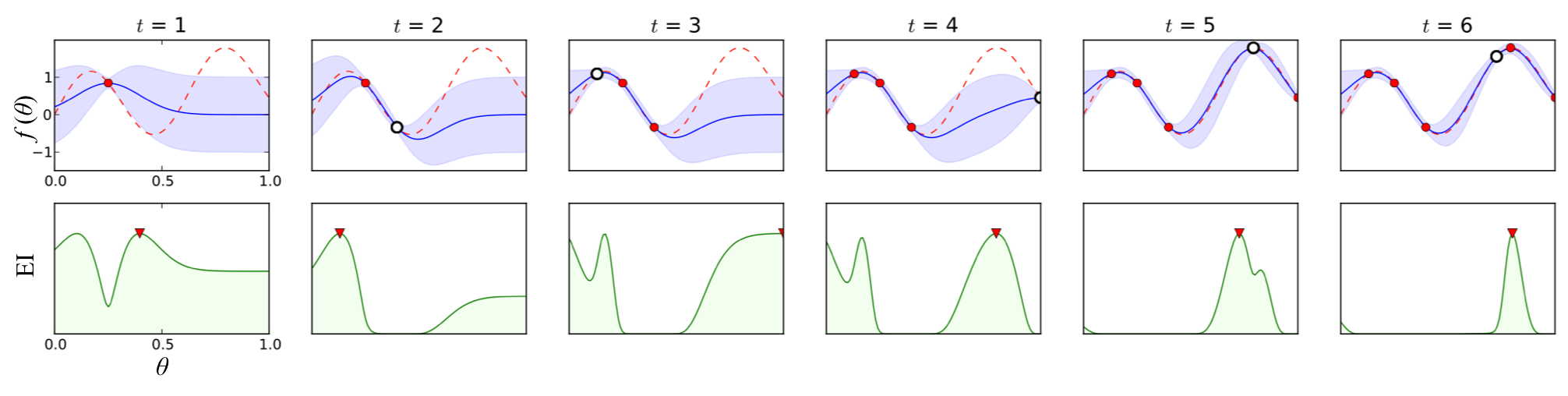}
\caption{One-dimensional illustration of Bayesian optimization with Gaussian processes (GPs) and the expected improvement acquisition (EI) function, over the first 6 iterations. The top plots show the GP mean in blue and the true unknown function in red. In the vicinity of query points, the uncertainty is reduced. The bottom plots shows the EI acquisition function and its proposed next query points. EI trades-off exploitation and exploration.} 
\label{fig:ei}
\end{figure}

Bayesian optimization is a sequential model-based approach to optimize black-box functions $f(\vtheta)$, $\vtheta \in \vTheta$. Its data efficiency makes it particularly suitable for expensive black-box evaluations like hyper-parameter evaluation.
Bayesian optimization specifies a probabilistic prior model over the unknown function $f$ and applies Bayesian inference to compute a
posterior distribution over $f$ given the previous observations $\{\vtheta_i, y_i\}_{i=1}^t$. This posterior distribution is in turn used to construct an acquisition function to decide the next query point $\vtheta_{t+1}$. The acquisition function trades-off exploitation and exploration.

One could use a wide variety of probabilistic models and acquisition functions, see for example the tutorial of \cite{shahriari2016taking}. In this work, we use Gaussian process (GP) priors over functions \citep{rasmussen2006gp} and the Expected Improvement (EI) acquisition function \citep{mockus1978ei}.
Figure \ref{fig:ei}  (from \cite{brochu2010tutorial}) illustrates Bayesian optimization with these choices in a 1D scenario.

The expected improvement at a given point $\vtheta$ is defined as
\begin{equation}
\mathrm{EI}(\vtheta) = \mathbb{E}_{f} [\max\{f(\vtheta) - f^\star, 0\}],
\end{equation}
where $f^\star$ is a target value, usually the best past observation or best posterior mean at past query points. Hence, $f^\star$ can be thought of as an aspiration value. EI attempts to do better than $f^\star$, but instead of greedily exploiting, it also uses the estimates of uncertainty derived by the probabilistic model over $f$ to explore the space $\vTheta$.

The choice of Bayesian optimization to tune the hyper-parameters of AlphaGo was motivated by several factors: (i) the win-rate $p(\vtheta)$ is not differentiable, (ii) large computational resources are required to evaluate the win-rate at a single hyper-parameter setting, and (iii) the number of hyper-parameters is moderate, making it possible to find a good setting within a few hundred steps.

We use a modified version of Spearmint \citep{Snoek:2012} with input warping to conduct Bayesian optimization. The hyper-parameter tuning procedure is summarized in Algorithm 1.

\begin{algorithm}
\caption{Algorithm to tune AlphaGo's hyper-parameters.}
\label{alg:tuning}
\begin{algorithmic}
\REQUIRE A baseline player with fixed parameters $\vtheta_0$
\FOR{$t = 1 \to T$}
\STATE Wait until computational resources becomes available
\STATE Incorporate recently finished self-play outcomes, if any, into the GP model
\STATE Propose a new value $\vtheta_t$ that maximizes the EI acquisition function
\STATE Launch a new set of self-play games with parameter $\vtheta_t$ against the baseline player.
\ENDFOR
\STATE Search for the final recommendation $\vtheta^*$ with the highest posterior mean of $p(\vtheta)$
\RETURN $\vtheta^*$.
\end{algorithmic}
\end{algorithm}

In black-box optimization, most algorithms assume the function value is observed either exactly or noisily with an unknown noise magnitude. In contrast, in self-play games, we can estimate the observation noise as the observed win-rate is an average of Bernoulli variables. When the number of games is large enough, the observed value is normally distributed according to the central limit theorem, and the noise standard deviation can be estimated as
\begin{equation}
\hat{\epsilon} = \sqrt{\frac{\bar{p} (1-\bar{p})}{N}}
\end{equation}
We adopted a Gaussian process model with a nonstationary Gaussian observation noise model. For every parameter setting, we supplied the Gaussian process model with the observed win-rate and the estimated standard deviation. The estimated $\hat{\epsilon}$ was clipped from below by $\sqrt{(N-1) / N^3}\approx 1/N$ to avoid ignoring the noise when observing all win/lose games.

We can reduce the cost of a function evaluation by decreasing the number of self-play games per hyper-parameter candidate at the price of higher noise. (While this is reminiscent of the general idea of early stopping in Bayesian optimization, it is simpler in our domain.)
Guided by this and the central limit theorem for Bernoulli random variables, we chose to evaluate $p(\vtheta)$ with $50$ games.

\begin{figure}[t]%
\centering
\begin{subfigure}[b]{0.75\textwidth}
\includegraphics[width=\textwidth]{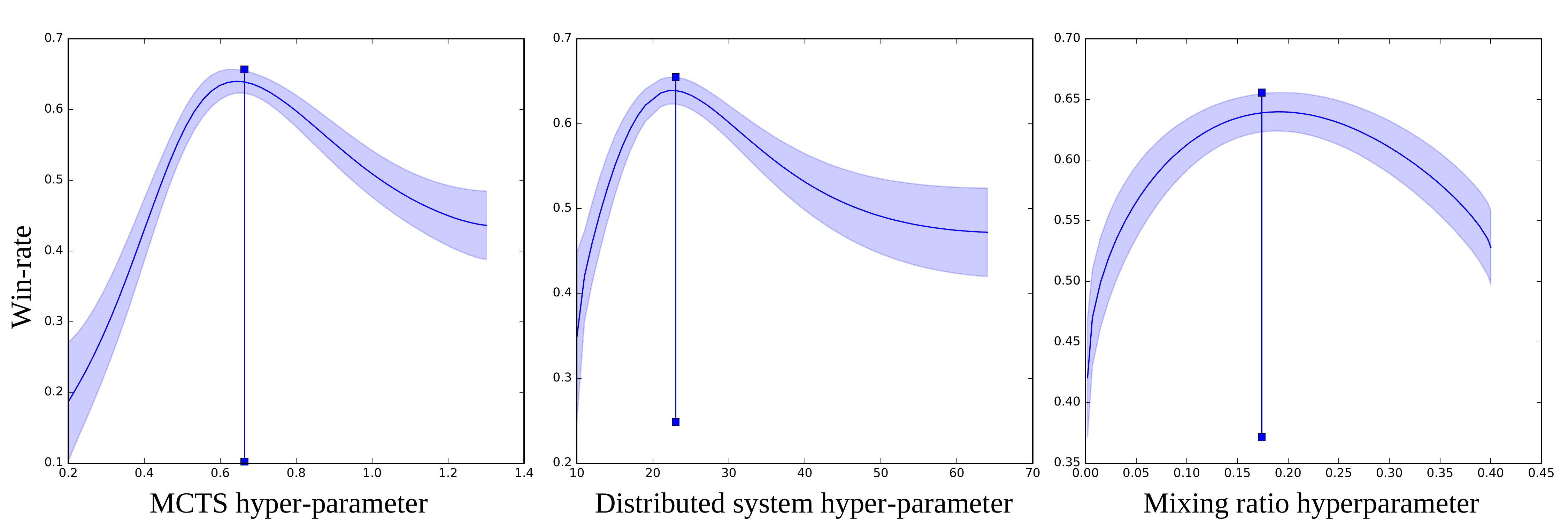}   
\label{fig:sensitivity_1d}
\end{subfigure}
\begin{subfigure}[b]{0.24\textwidth}
\includegraphics[width=\textwidth]{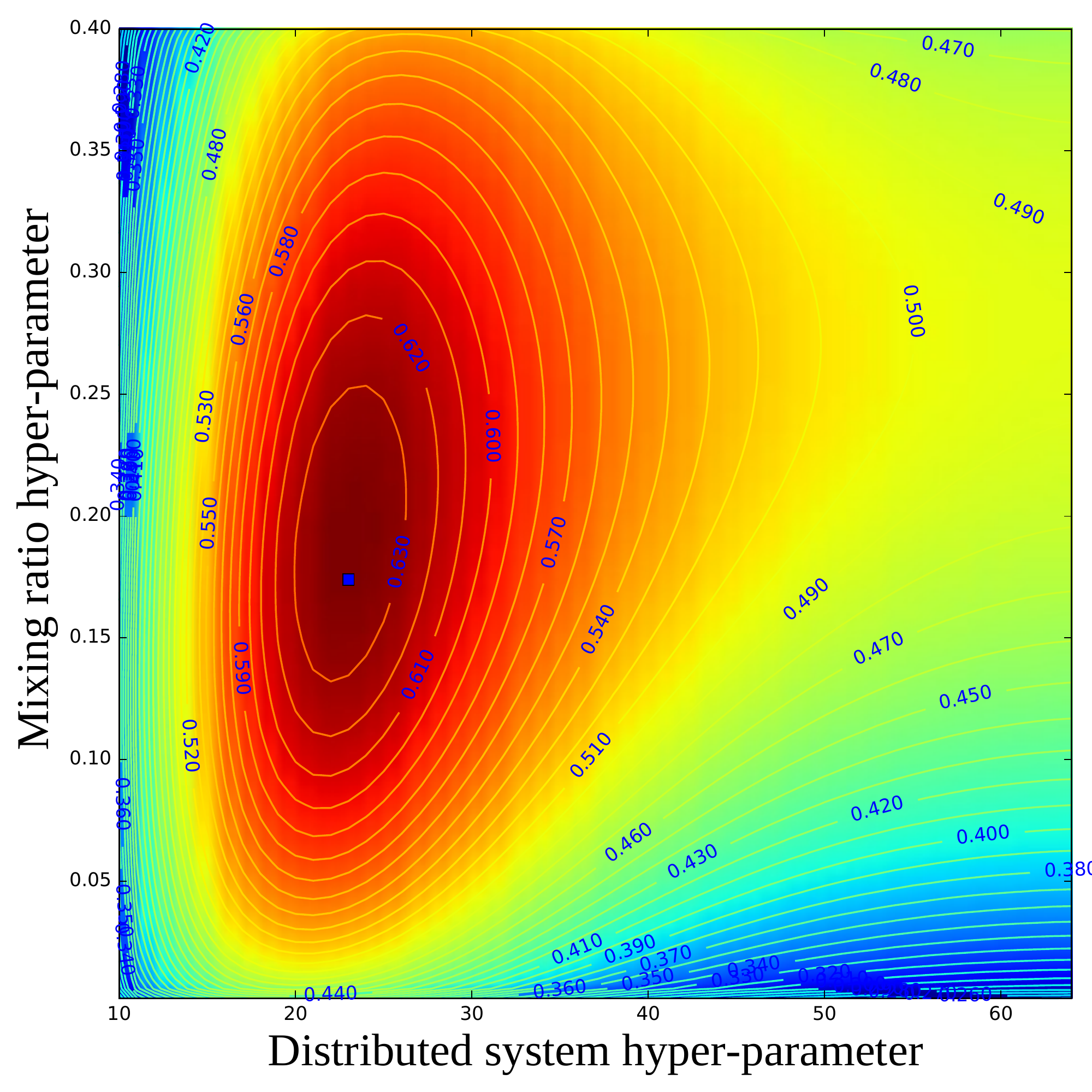}
\label{fig:sensitivity_2d}
\end{subfigure}
\caption{Leftmost three plots: estimated posterior mean and variance of the win-rate for three individual hyper-parameters while fixing the remaining hyper-parameters. The vertical bar shows the fixed reference parameter value. Rightmost plot: posterior mean for two hyper-parameters, showing the correlation among these.}
\label{fig:sensitivity}
\end{figure}

Finally, we developed a visualization tool to understand the win-rate sensitivity with respect to each individual hyper-parameter. Specifically, we plot the win-rate posterior mean and variance as a function of one or a pair of hyper-parameters while holding the other hyper-parameters fixed, as illustrated in Figure~\ref{fig:sensitivity}. We also estimated the contribution of each parameter to the difference in playing strength between two settings. We found it useful to understand the importance of each hyper-parameter and the correlations among the hyper-parameters. We used this information to select the most influential hyper-parameters for optimization in subsequent AlphaGo versions.

\section{Tasks and Results}

In the following subsections, we describe the various tasks where Bayesian optimization was applied, and found to yield fruitful results.

\subsection{Task 1: Tuning MCTS hyper-parameters}
\label{sec:task1}

We optmized the MCTS hyper-parameters governing the UCT exploration formula \citep[Section Search]{Silver:2018}, network output tempering, and the mixing ratio between the fast roll-out value and value network output. The number of hyper-parameters to tune varied from 3 to 10. 

\begin{figure}[t!]%
\centering
\includegraphics[width=0.7\textwidth]{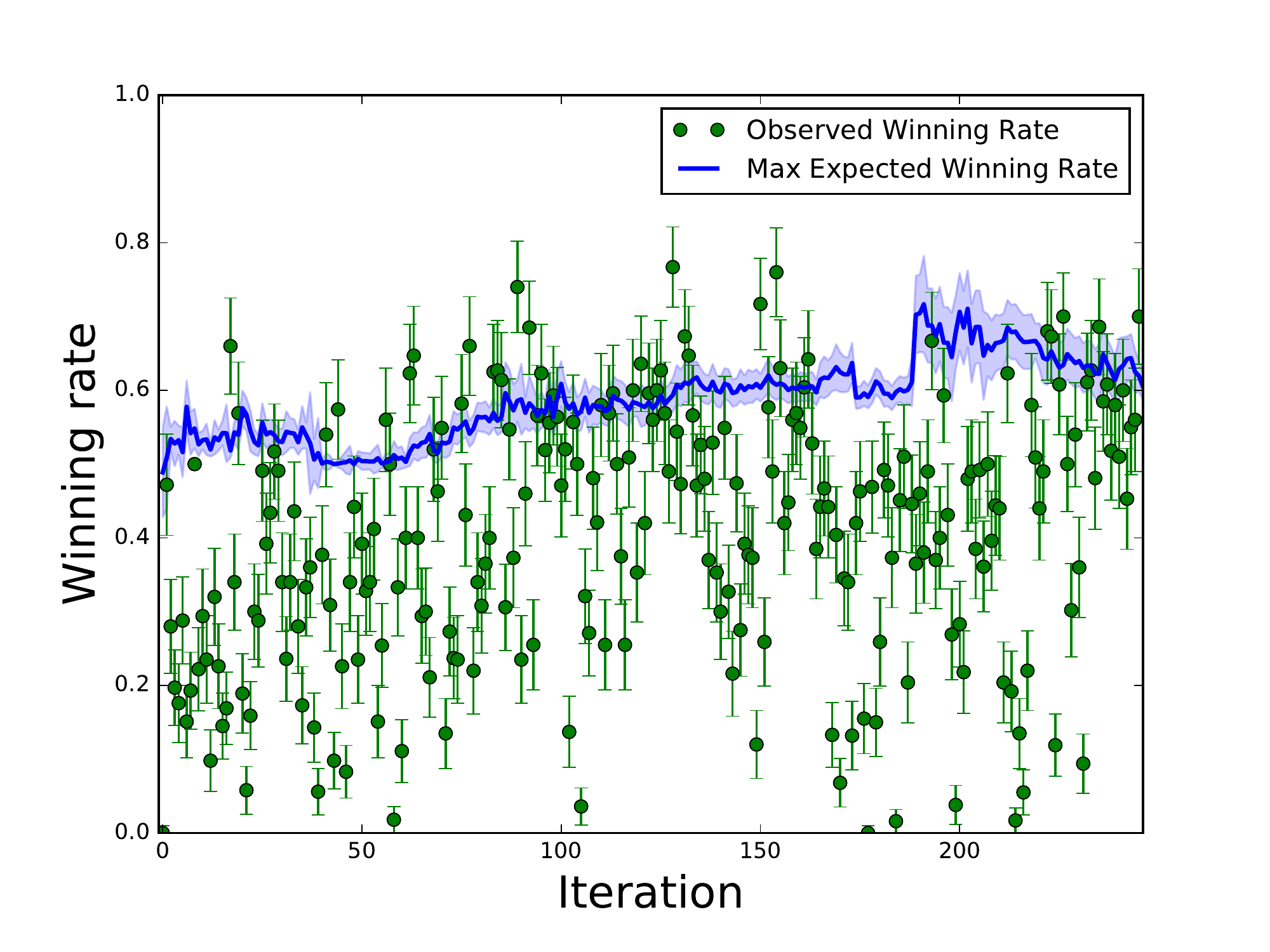}
\caption{Typical values of the observed and maximum expected win-rates as a function of the optimization steps.}
\label{fig:tuning_progress}
\end{figure}

The development of AlphaGo involved many design iterations.
After completing the development of an AlphaGo version, we refined it with Bayesian optimization and self-play. At the start of each design iteration, the win-rate was 50\%. However, by tuning the MCTS hyper-parameters this win-rate increased to 63.2\% and 64.4\% (that is, 94 and 103 Elo gains) in two design iterations prior to the match with Lee Sedol. Importantly, every time we tuned a version, the gained knowledge, including hyper-parameter values, was passed on to the team developing the next version of AlphaGo. That is, the improvements from all tuning tasks were compounded. After the match with Lee Sedol, we continued optimizing the MCTS hyper-parameters, resulting in progressively stronger AlphaGo agents.
Figure \ref{fig:tuning_progress} shows a typical curve of the win-rate against an opponent of the same version with fixed hyper-parameters.
 
 Interestingly, the automatically found hyper-parameter values were very different from the default values found by previous hand tuning efforts.
Moreover, the hyper-parameters were often correlated, and hence the values found by Bayesian optimization were not reachable with element-wise hand-tuning, or even by tuning pairs of parameters in some cases.

By tuning the mixing ratio between roll-out estimates and value network estimates, we found out that Bayesian optimization gave increased preference to value network estimates as the design cycle progressed. This eventual led the team to abandon roll-out estimates in future versions of AlphaGo and AlphaGo Zero \citep{Silver:2017}.

\subsection{Task 2: Tuning fast AlphaGo players for data generation}
\label{sec:task2}

We generated training datasets for the policy and value networks by running self-play games with a very short search time, \emph{e.g.,} $0.25$ seconds in contrast to the regular search time. The improvement of AlphaGo over various versions depended on the quality of these datasets. Therefore, it was crucial for the fast players for data generation to be as strong as possible. Under this special time setting, the optimal hyper-parameters values were very different, making manual tuning prohibitive without proper prior knowledge. Tuning the different versions of the fast players resulted in Elo gains of 300, 285, 145, and 129 for four key versions of these players.

\subsection{Task 3: Tuning on TPUs}
\label{sec:task3}

Tensor Processing Units (TPUs) provided faster network evaluation than GPUs. After migrating to the new hardware, AlphaGo's performance was boosted by a large margin. This however changed the optimal value of existing hyper-parameters and new hyper-parameters also arose in the distributed TPU implementation. Bayesian optimization yielded further large Elo improvements in the early TPU implementations.

\subsection{Task 4: Developing and tuning a dynamic mixing ratio formula}
\label{sec:task4}

Early versions of AlphaGo used a constant mixing ratio between the fast roll-out and value network board evaluation, regardless of the stage of a game and the search time. This was clearly a sub-optimal choice, but we lacked proper technique to search for the optimal mixing function. With the introduction of Bayesian optimization, we could define a more flexible formula and search for the best formula parameters. In particular, we defined the new dynamic mixing ratio $r$ at a tree node as a function of the move number, $m$, and number of node visits, $n$, during tree search in the following form:
\begin{equation}
r(m, n) = \sigma(\sigma^{-1}(r_0) + a m + b \log n)
\end{equation}
where $\sigma(x)=1/(1+\exp(-x))$ is the logistic function to restrict $r \in (0, 1)$, $r_0$ is the parameter corresponding to the mixing ratio at move 0 without search, and $a$ and $b$ are linear coefficients.

\begin{figure}[t]
\centering
\begin{subfigure}[t]{0.32\textwidth}
\includegraphics[width=\textwidth]{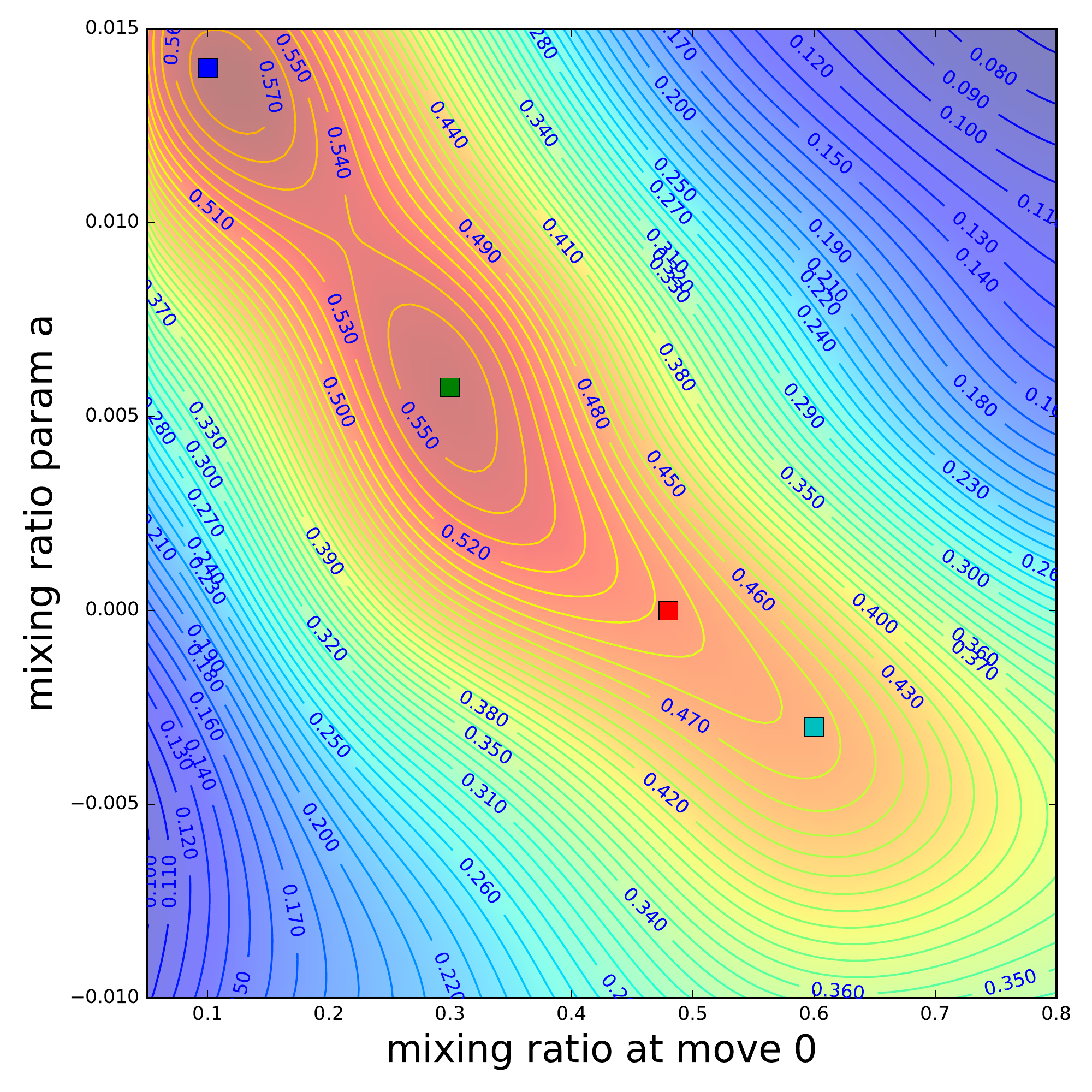}
\caption{Correlation between $r_0$ and $a$}
\label{fig:dynamic_mixing_ratio_0}
\end{subfigure}
\begin{subfigure}[t]{0.32\textwidth}
\includegraphics[width=\textwidth]{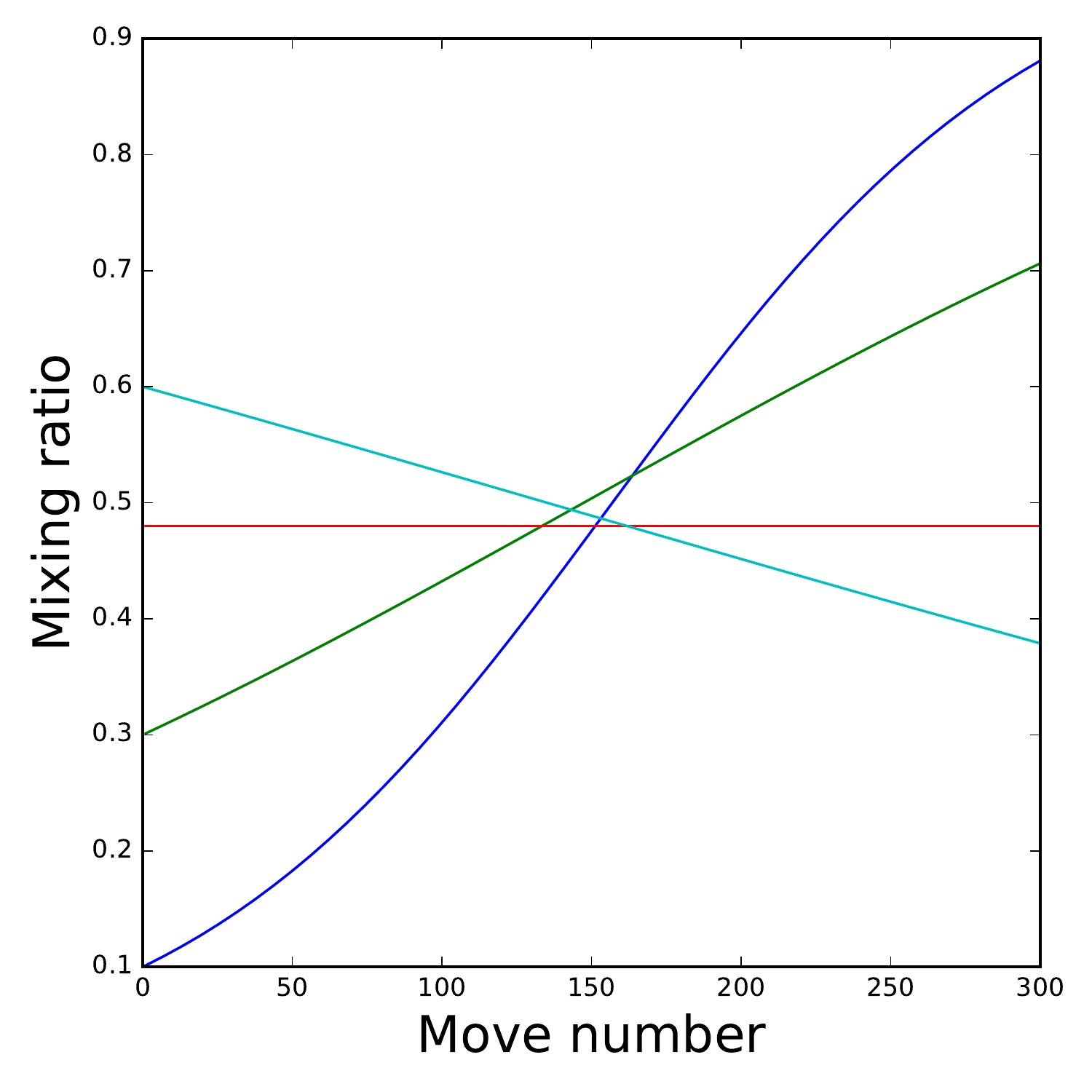}
\caption{Mixing ratio along the ridge}
\label{fig:dynamic_mixing_ratio_curves}
\end{subfigure}
\begin{subfigure}[t]{0.32\textwidth}
\includegraphics[width=\textwidth]{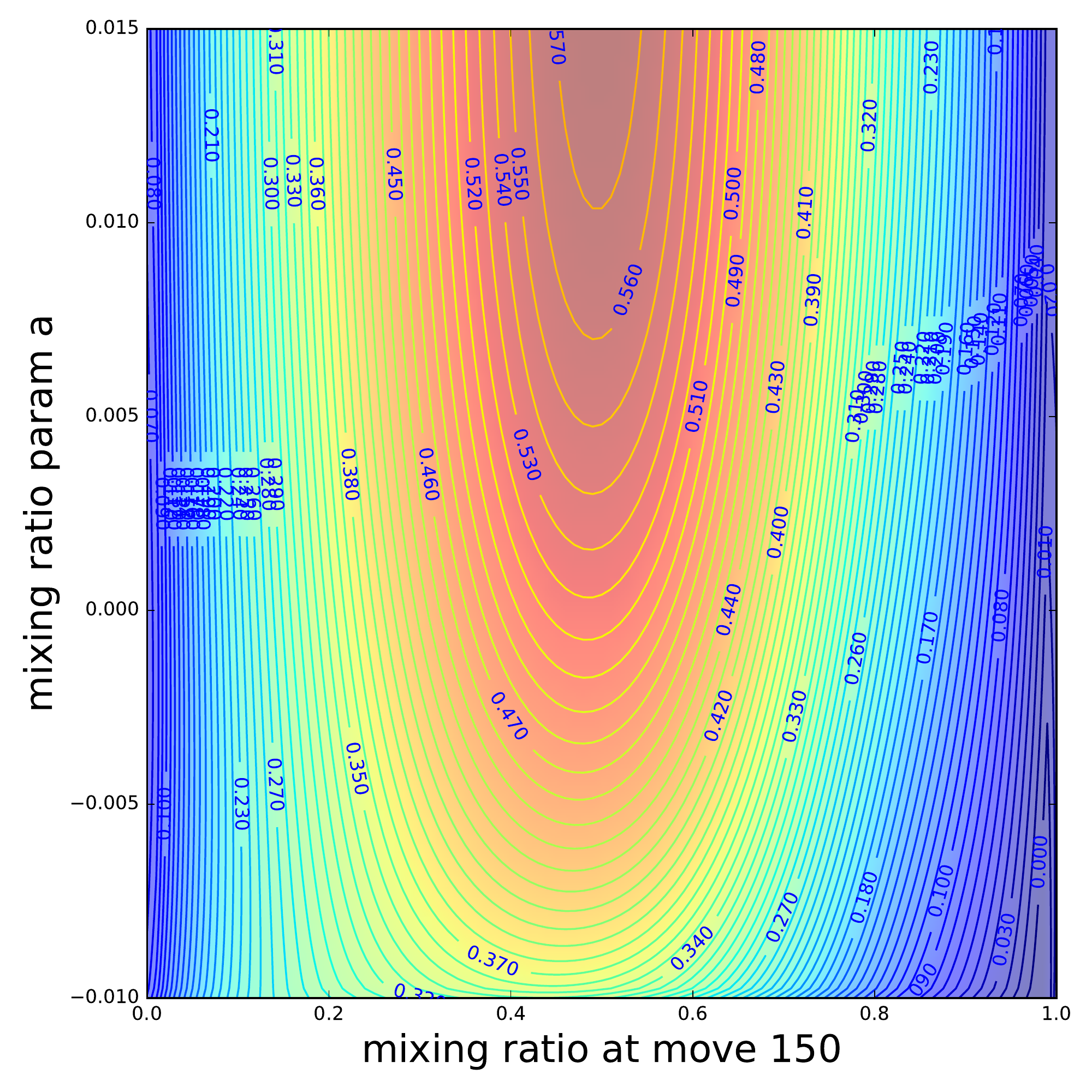}
\caption{Correlation between $r_{150}$ and $a$}
\label{fig:dynamic_mixing_ratio_150}
\end{subfigure}
\caption{Tuning the dynamic mixing ratio formula of AlphaGo.}
\label{fig:dynamic_mixing_ratio}
\end{figure}

\begin{figure}[t]
\centering
\includegraphics[width=\textwidth]{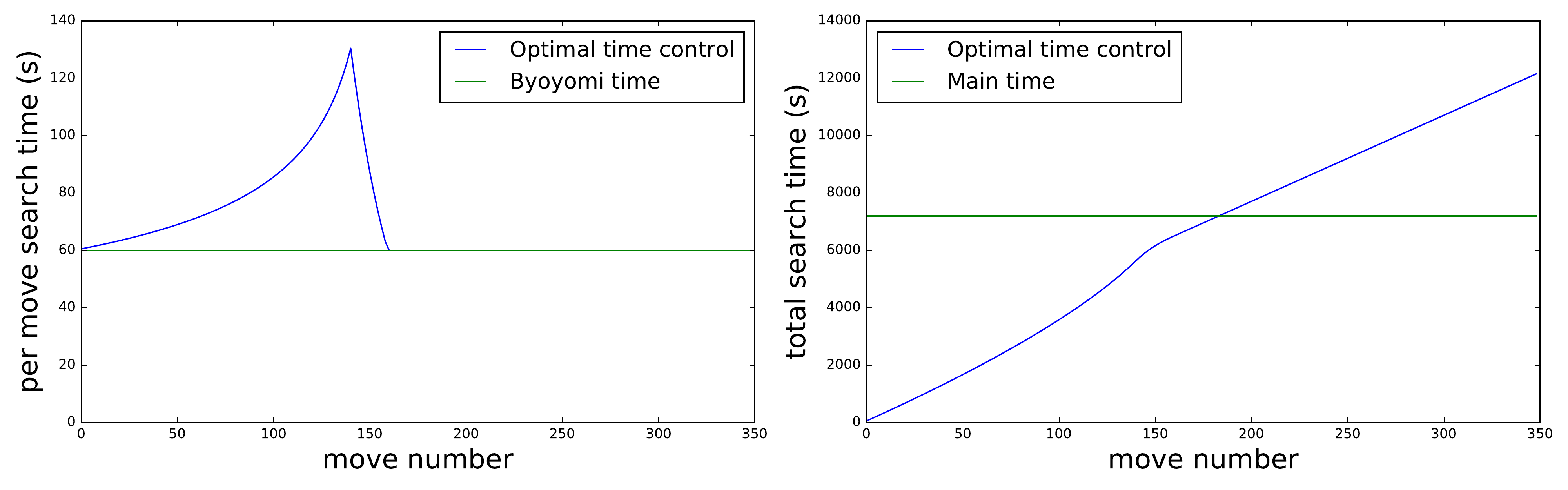}
\caption{Optimized time control formula for AlphaGo with a 2-hour main play time and 60-second byoyomi.}
\label{fig:time-control}
\end{figure}

During optimization, we observed that the best value of $r_0$ was highly correlated with $a$ as illustrated in Figure \ref{fig:dynamic_mixing_ratio_0}. This caused difficulties in optimizing the parameters jointly and resulted in failed hand-tuning attempts. By inspecting the ridge of high win-rate in the figure, we noticed that when $b=0$, all the points along the ridge corresponded to a mixing ratio formula that crossed around the point $(m=150, r=0.48)$. Figure \ref{fig:dynamic_mixing_ratio_curves} shows four mixing ratio vs.\ move number curves corresponding to the four points in plot \subref{fig:dynamic_mixing_ratio_curves}. This suggested that it was important to find a good value for the mixing ratio at around move 150. This finding was consistent with the observation that the game-determining-points in AlphaGo's self-play games usually happened between moves 150 and 200.

With this observation, we reparameterized the mixing formula as
\begin{equation}
r(m, n) = \sigma(\sigma^{-1}(r_{150}) + a (m - 150) + b \log n)
\end{equation}
with $r_{150}$ indicating the mixing ratio at move 150, and obtained an uncorrelated function as shown in Figure \ref{fig:dynamic_mixing_ratio_150}. The optimal dynamic mixing ratio formula is the green curve in Figure \ref{fig:dynamic_mixing_ratio_curves}. With this formula, AlphaGo placed more weight on value networks at the beginning and less towards the end of a game. This was consistent with the facts that the value network was better at global opening judgment than roll-outs, and that roll-outs became accurate in end games when search depth reduced.

The parameter $b$ was also highly correlated with $r_0$ and we did find a reparameterization form to decorrelate them, but the optimal value for $b$ turned out to be 0 after tuning, suggesting that we do not need to include the visit number in the mixing ratio formula.

\subsection{Task 5: Tuning a time control formula}
\label{sec:task5}

MCTS is an anytime algorithm, whose tree search can be interrupted at any point, returning to the current best choice. To prepare for the formal match with Lee Sedol, which had a main time of 2 hours and 3 60-second byoyomi per player, we wanted to optimize the search time allocation across all moves. We considered time allocation as an optimization problem, so as to maximize the win-rate of a player subject to the time restriction against another one with a fixed time schedule. \citet{huang2010time} proposed a parameterized time control formula to allocate search time for a fixed main time without byoyomi. We adopted a more flexible formula including the byoyomi time, as follows
\begin{align}
t(m|a, b, c, m_{\mathrm{peak}}) &= \max\left\{
a \frac{t_{\mathrm{remain}}(m)}{\max\{m_{\mathrm{peak}} - m, 0\} + b},
f(m) t_{\mathrm{byo}}\right\}, \\
t_{\min}(m) &= \left\{ 
\begin{array}{ll}
c , & \textrm{ if } m \leq \mathrm{peak} \\
1, & \textrm{ otherwise}
\end{array} 
\right. ,
\end{align}
where $m$ is the 1-based move number, $m_{\mathrm{peak}} \in [1, 250]$ is the move with the peak search time, $t_{\mathrm{remain}}(m) := t_{\mathrm{main}} - \sum_{\tau=1, m-1} m(\tau)$ is the remaining main time at move $m$, $t_{\mathrm{byo}}$ is the byoyomi time, and $\mathbb{I}[x] = 1$ if $x$ is true, otherwise 0.  $a \in [0, 2)$, $b > 1$, $c \in [0, 1]$ are hyper-parameters to tune together with $m_{\mathrm{peak}}$. The formula reduces to that in \cite{huang2010time} when $a=1$ and $t_{\mathrm{byo}}=0$.

The optimal formula after tuning all the hyper-parameters is shown in Figure \ref{fig:time-control}. AlphaGo obtains an improvement of a 66.5\% win-rate against the default time setting with a fixed 30-second search time per move. Interestingly, the move with the peak search time under the optimal time control formula is also around move 150.

\section{Conclusion}

Bayesian optimization provided an automatic solution to tune the game-playing hyper-parameters of AlphaGo. This would have been impossible with traditional hand-tuning. Bayesian optimization contributed significantly to the win-rate of AlphaGo, and helped us gain important insights which continue to be instrumental in the development of new versions of self-play agents with MCTS.

\section{Acknowledgments}

We are very thankful to Misha Denil and Alexander Novikov for providing us with valuable feedback in the preparation of this document.

\bibliographystyle{plainnat}
\bibliography{alphago}

\end{document}